\documentclass[sigconf]{acmart}

\usepackage{booktabs} 

\usepackage{float}
\usepackage{times}
\usepackage{latexsym}
\usepackage{graphicx}
\usepackage{url}
\usepackage{hyperref}
\usepackage{amsmath}
\usepackage{comment}
\usepackage{makecell}
\usepackage{multirow}
\usepackage{subcaption}
\usepackage{caption}
\usepackage[title]{appendix}
\usepackage{tabularx}
\usepackage{booktabs}
\usepackage{longtable}
\setcopyright{acmcopyright}

\acmArticle{4}
\acmPrice{15.00}

\begin{document}
\title{\textsc{AutoLLM-Card}: Towards a Description and Landscape of Large Language Models}
  
\renewcommand{\shorttitle}{SIG Proceedings Paper in LaTeX Format}

\author{Shengwei Tian}
\orcid{1234-5678-9012}
\affiliation{%
  \institution{UoM}
  \streetaddress{Oxford Rd}
  \city{Manchester} 
  \state{England} 
  \country{UK}
  \postcode{M13}  
}
\email{shengwei.tian@postgrad.manchester.ac.uk}

\author{Lifeng Han}
\orcid{1234-5678-9012}
\affiliation{%
  \institution{UoM, Mancheter, UK}
  \city{LUMC and LIACS, Leiden, NL} 
  \state{} 
  \country{}
  \postcode{M13}  
}
\email{l.han@lumc.nl}

\author{Goran Nenadic}
\orcid{1234-5678-9012}
\affiliation{%
  \institution{UoM}
  \city{Manchester} 
  \state{England} 
  \country{UK}
  \postcode{M13}  
}
\email{goran.nenadic@manchester.ac.uk}

\begin{abstract}
With the rapid growth of the Natural Language Processing (NLP) field, a vast variety of Large Language Models (LLMs) continue to emerge for diverse NLP tasks. As more papers are published, researchers and developers face the challenge of information overload. Thus, developing a system that can automatically extract and organise key information about LLMs from academic papers is particularly important. The standard format for documenting information about LLMs is the LLM model card (\textbf{LLM-Card}). 
We propose a method for automatically generating LLM model cards from scientific publications. 
We use Named Entity Recognition (\textbf{NER}) and Relation Extraction (\textbf{RE}) methods that automatically extract key information about LLMs from the papers, helping researchers to access information about LLMs efficiently. These features include model \textit{licence}, model \textit{name}, and model \textit{application}. With these features, we can form a model card for each paper. 
We processed 106 academic papers by defining three dictionaries -- LLM's name, licence, and application. 
11,051 sentences were extracted through dictionary lookup, and the dataset was constructed through manual review of the final selection of 129 sentences with a link between the name and the \textit{licence}, and 106 sentences with a link between the model name and the \textit{application}.
The resulting resource is relevant for LLM card illustrations using relational knowledge graphs. 
Our code and findings can contribute to automatic LLM card generation.
Data and code in \textsc{autoLLM-Card} will be shared and freely available at \url{https://github.com/shengwei-tian/dependency-parser-visualization}
\end{abstract}

%
%



\maketitle

\section{Introduction}

As a major approach in natural language processing (NLP), language modelling has been widely studied for language understanding and generation in the past two decades, evolving from statistical language models to neural language models. Recently, Pre-trained Language Models (PLMs) have been proposed by pretraining Transformer models over large-scale corpora, showing strong capabilities in solving various NLP tasks. 


In recent years, major research institutes and technology companies have increased their investment in the development of LLMs, resulting in a series of groundbreaking models with the potential for a wide range of applications. OpenAI is a pioneer in this field. 
The GPT family (e.g., GPT-3 and GPT-4) has made significant advances in natural language generation, dialogue systems, and text comprehension, and is widely used in a variety of intelligent assistants, content creation tools, and automated services~\cite{torfi2020natural}. Meanwhile, the Bidirectional Encoder Representations from Transformers (BERT)~\cite{devlin-etal-2019-bert} and the subsequent T5 (Text-to-Text Transfer Transformer)~\cite{raffel2020exploring} models developed by Google research department have also made significant progress in tasks such as question-answer systems, text classification and information retrieval. 
The Roberta and BlenderBot series of models from Facebook (now Meta) have demonstrated powerful performance in social media data analysis and conversational AI~\cite{liu2020roberta}.

NLP technologies, driven by Large Language Models (LLMs), have skyrocketed and revolutionised the field of NLP, leading to breakthroughs in various tasks such as machine translation, text summarisation, and dialogue systems. However, these advances have been accompanied by the challenge of navigating the \textbf{vast array} of recent research findings. Each year, thousands of papers on LLMs are published, presenting new models, methods, and innovations. This leads to a phenomenon of information \textbf{overload} that makes it difficult for even the most experienced researchers to \textit{keep up with the latest developments}.

\begin{figure}[th]
    \centering
    \includegraphics[width=0.99\linewidth]{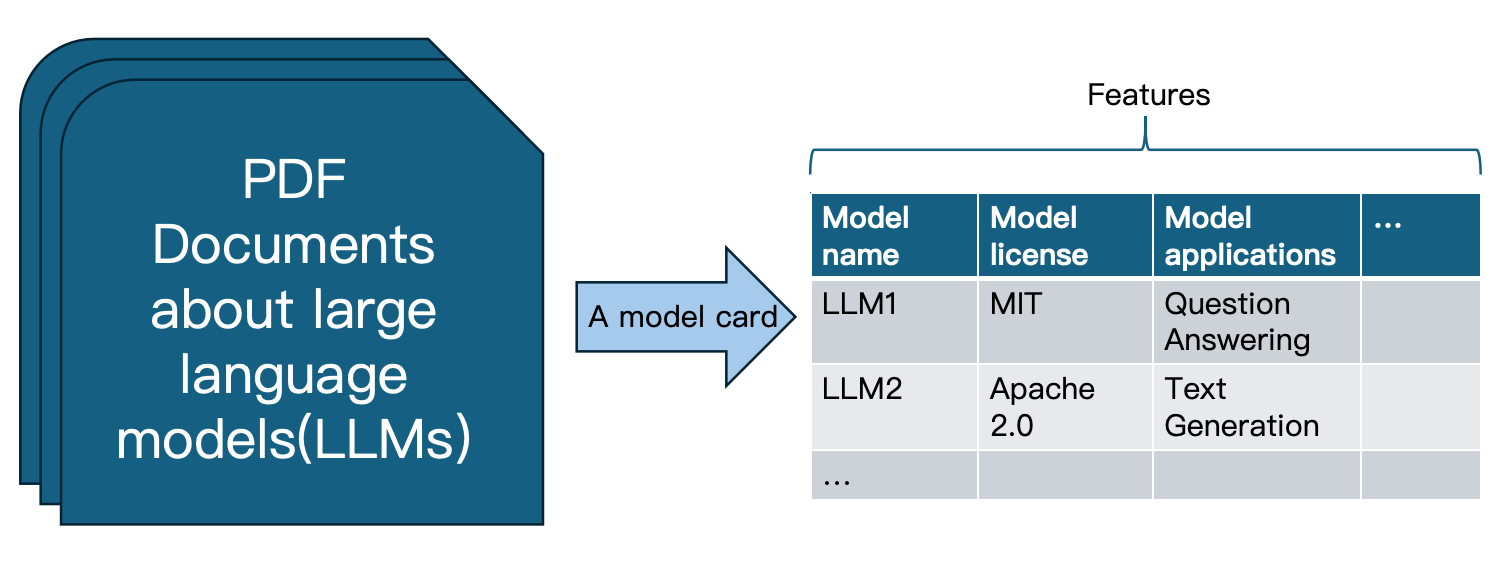}
    \caption{\textsc{autoLLM-card} method demonstration}
    \label{Demonstraiton}
\end{figure}

We need a convenient and efficient tool that enables researchers and developers to quickly sift through and understand large numbers of models and their properties, i.e. the \textsc{\textbf{LLM model-card}}. 
Model cards often contain the metadata of the model providing some detailed technical documentation such as model parameters, usage, license, etc. \cite{singh2023unlocking,liu-etal-2024-automatic} \footnote{\url{https://huggingface.co/docs/hub/en/model-cards}}.
Motivated by these needs, this work \textit{proposes} a novel system that automatically extracts such relationships, as illustrated in Figure~\ref{Demonstraiton}, the \textsc{AutoLLM-Card}. Our goal is to simplify the process of understanding key information about \textit{licence} types, \textit{application} areas, and more for researchers. By automating this process, researchers can gain a clearer understanding of the latest models, allowing them to save time to focus on innovation rather than tediously sifting through extensive literature.

This work mainly focuses on 3 Research Questions (RQs):
\begin{itemize}
    \item RQ1: How to \textbf{identify} the target sentences that are related to the description of LLMs? 
    \item RQ2: How to \textbf{model} the relationships between LLMs and their licences, and model relationships between LLMs and their applications.
    \item RQ3:  How to address the issue of \textbf{data scarcity} for tasks where no readily available dataset exists for LLMs descriptions. 
    \end{itemize}

\section{Related work} 
\label{literature_review}
In this section, we talk about LLMs, Named Entity Recognition (NER), Relation Extraction (RE), and Model card generations.

\subsection{Large Language Models}
\label{llms}

 
A pivotal moment in the history of LLMs came with the development of word \textit{embeddings}, such as word2vec  \cite{mikolov2013efficient}, which enabled models to learn distributed representations of words in a continuous vector space. This innovation laid the groundwork for more complex architectures like Recurrent Neural Networks (RNNs) and Long Short-Term Memory (LSTM) networks, which were among the first to effectively model sequential data, capturing dependencies across different points in a text sequence.
 

Following BERT, the field of NLP has progressed \textbf{rapidly} with a series of more powerful models, notably OpenAI's GPT family. GPT-2, released in 2019, significantly improves AI text generation and closes the gap with human writing. In 2020, GPT-3 set a new benchmark with 175 billion parameters for performing small or no samples for complex tasks such as code writing and dialogue generation \cite{brown2020language}. In 2022, ChatGPT based on GPT-3.5 further optimises dialogue performance for more natural interactions.
GPT-4 in 2023 extends to multi-modal tasks, handling text and image inputs, showing potential for a wide range of applications in fields such as healthcare and law \cite{openai2023gpt}.

However, as the \textbf{scale} of the models and the range of \textbf{applications} have increased, LLMs have also sparked discussions about \textit{interpretability}, \textit{safety}, and \textit{ethical} issues. While these models drive changes in technology and human interaction, their potential risks and limitations need to be treated with caution to ensure that social fairness and safety are maintained while driving innovation.

\subsection{Named Entity Recognition}




Named Entity Recognition (\textbf{NER}) is a subtask of information extraction in NLP, which is used to classify named entities into predefined categories such as names of people, organisations, places, etc. In the field of NLP, understanding these entities is crucial for many applications because they often contain the most important information in a text. NER is a bridge between \textit{unstructured} text and \textit{structured} data, enabling machines to sift through large amounts of textual information and extract valuable data in the form of classification. By precisely locating specific entities in a sea of words, NER changes the way we process and utilise textual data. In text mining, a named entity is a word or a phrase that clearly identifies an entity, an entity with similar properties, from a set of other items with similar properties. In the formulation of named entities, the word named limits the scope of entities that have one or more rigid designators that represent the referent. Typically, rigid designators often include proprietary names, but this depends on the domain of interest, which has reference words for objects as named entities \cite{sharnagat2014named}. 
For example, in the sentence: BERT was released under the Apache licence 2.0, the entity of \textbf{licence} is Apache licence and the \textbf{entity} of the model name is BERT. 

\begin{figure}[th]
    \centering
    \includegraphics[width=0.97\linewidth]{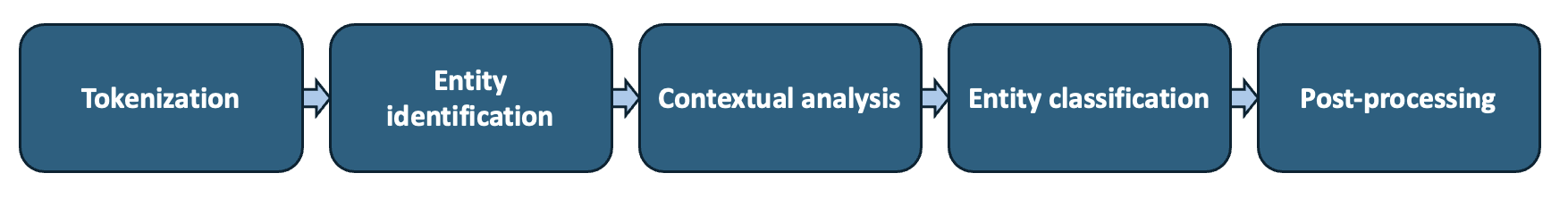}
    \caption{The steps for named entities recognition}
    \label{The-steps-of-NER}
\end{figure}

As shown in Figure \ref{The-steps-of-NER},  NER can be broken down into several steps: Tokenisation before identifying entities, the text is split into tokens, which can be words, phrases, or even sentences. 
Entity identification involves recognising items such as capitalisation in licence (`Apache 2.0’) or specific formats. Once entities are identified, they are grouped into predefined categories such as `\textbf{model name}', `\textbf{licence}' or `\textbf{application}' in our task. 
This is typically achieved by machine learning models trained on annotated datasets. In the example ``BERT was released under the Apache licence 2.0'', `BERT' would be categorised as `model name', while `Apache 2.0’ would be categorised as `licence'. The NER system usually considers the surrounding context to improve accuracy. After the initial identification and classification, \textit{post-processing} may be performed to refine the results, which may involve resolving ambiguities, merging multi-tagged entities, or augmenting the entity data with a knowledge base \cite{jurafsky2000speech}.

\subsection{Relation Extraction}

\subsubsection{Rule-based methods}

\begin{figure}[th]
    \centering
    \includegraphics[width=0.99\linewidth]{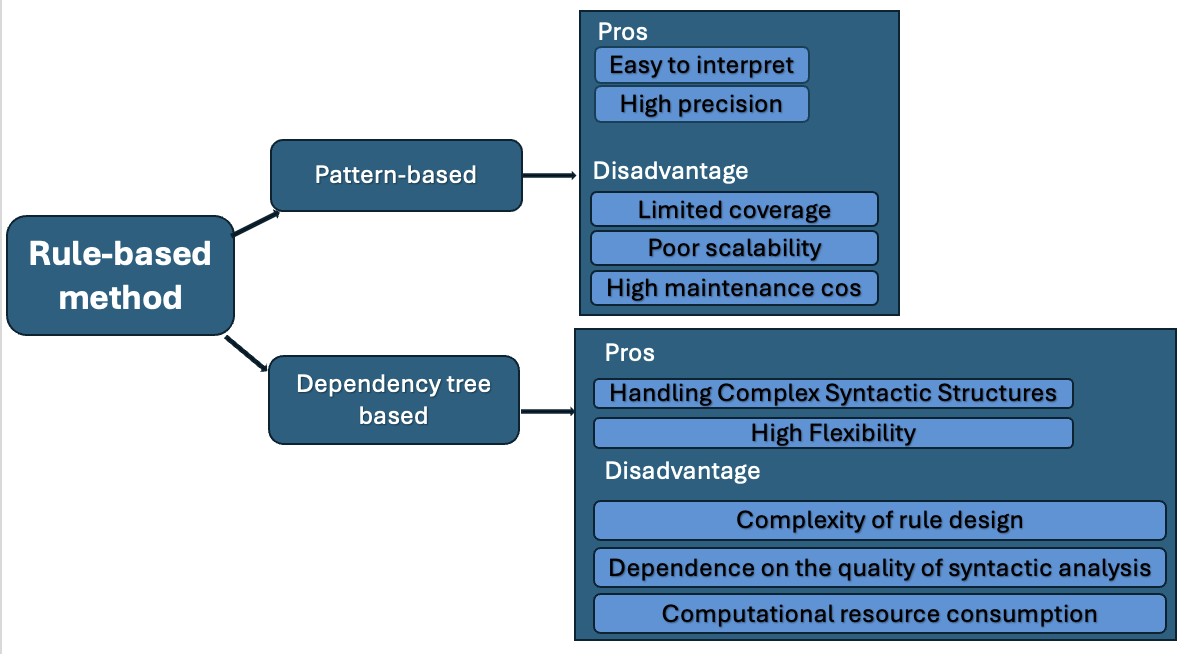}
    \caption{Rule-based methods for Relation Extractions}
    \label{Rule-based}
\end{figure}

The rule-based relation extraction approach as shown in Figure \ref{Rule-based} is one of the early techniques in the field of NLP, which relies on linguists and domain experts to manually write rules to extract relations from text based on an in-depth understanding of linguistic structures \cite{zhang2009rule}. The core of this approach lies in the use of \textit{predefined linguistic patterns} or syntactic structures to accurately match entities and their relationships in a text. While this approach excels in some specific domains and tasks, its scalability and generality are limited.

\textbf{Pattern-based methods}

Pattern-based relational extraction methods rely heavily on patterns written by linguistic experts, which are usually constructed from specific linguistic rules or \textit{regular expressions} to capture fixed relational expressions in the text. This approach is very \textit{effective} in \textbf{domain-specific} tasks, especially when the expected form of the relationship expression is relatively fixed. However, as the diversity of texts increases, the applicability of the fixed model begins to decline. For example, the sentences `France's capital is Paris' or `Paris, the capital of France', which express the same relation, both require different models to be captured \cite{sarawagi2008information}.
To cope with this linguistic diversity, pattern-based approaches usually require the design of a large number of rules to cover the possible linguistic variants. These rules can be generated either by hand or by using some semi-automated tools, but either way, writing and maintaining them is very time-consuming and labour-intensive.

\textbf{Advantages:}
\begin{itemize}
\item \textbf{High precision}: In certain highly structured domains, pattern-based methods can achieve very high precision. It is possible to extract relations precisely and directly for a specific sentence pattern or expression.
\item \textbf{Easy to interpret}: Since the rules are hand-written, each extracted relation has a clear logical basis and is easy for humans to understand and verify.
\end{itemize}

\textbf{Disadvantages:}
\begin{itemize}
 \item \textbf{Limited coverage}: Pattern-based rules can often only handle a limited number of sentence types, making it difficult to cope with the diversity and complexity of natural language, complex compound sentences or implicit relations often can not be extracted by simple patterns.
\item \textbf{Poor scalability}: Rules need to be redesigned and adapted when confronted with new textual domains or languages, which makes it difficult to scale pattern-based approaches to different domains or large-scale corpora.
\item \textbf{High maintenance cost}: Languages and expressions may change over time, which requires constant updating and maintenance of the rule base, increasing the cost of long-term use.
\end{itemize}

\textbf{Dependency tree-based methods}

Dependency Tree Based Relation Extraction method extracts more complex relations from sentences by analysing their dependency trees. The dependency tree is a commonly used syntactic structure representation in NLP, which represents the syntactic dependencies between words in a sentence. The advantage of the dependency tree approach is that it can handle \textbf{complex} grammatical structures, including long-distance dependencies and \textbf{nested} relationships. This ability makes the dependency tree-based approach outperform the simple pattern-based approach in processing complex sentences. Dependency tree-based rules usually involve multiple levels of conditions, e.g., the subject of a certain verb node should be a noun of a particular type, and the object of that verb needs to satisfy specific conditions. This fine-grained rule design allows the dependency tree-based approach to capture more complex types of relationships \cite{sachan2020syntax}.

\textbf{Advantages:}
\begin{itemize}
\item \textbf{Handling Complex Syntactic Structures}: Dependency trees are able to capture complex relationships between words in a sentence, allowing the method to handle long-distance dependencies and nested structures, which is useful for relational extraction of complex sentences \cite{tian2022improving}.
\item \textbf{High Flexibility}: The Dependency Tree method is able to adapt to a wide range of syntactic structures and cover a wide range of topics, allowing it to maintain a high accuracy rate in diverse texts.
\end{itemize}

\textbf{Disadvantages:}
\begin{itemize}
\item \textbf{The complexity of rule design}: Due to the complexity of dependency trees, designing effective rules requires deep linguistic knowledge and a lot of experiments, making rule writing difficult and costly.
\item \textbf{Dependence on the quality of syntactic analysis}: The accuracy of the dependency tree depends on the performance of the syntactic analyser; if the syntactic analyser produces wrong dependencies, the result of the relation extraction will also be affected.
\item \textbf{Computational resource consumption}: The construction and parsing of dependency trees usually require high computational resources, which limits their application in large-scale real-time processing tasks.
\end{itemize}

\subsubsection{Deep Learning-based methods}

\begin{figure}
    \centering
    \includegraphics[width=0.99\linewidth]{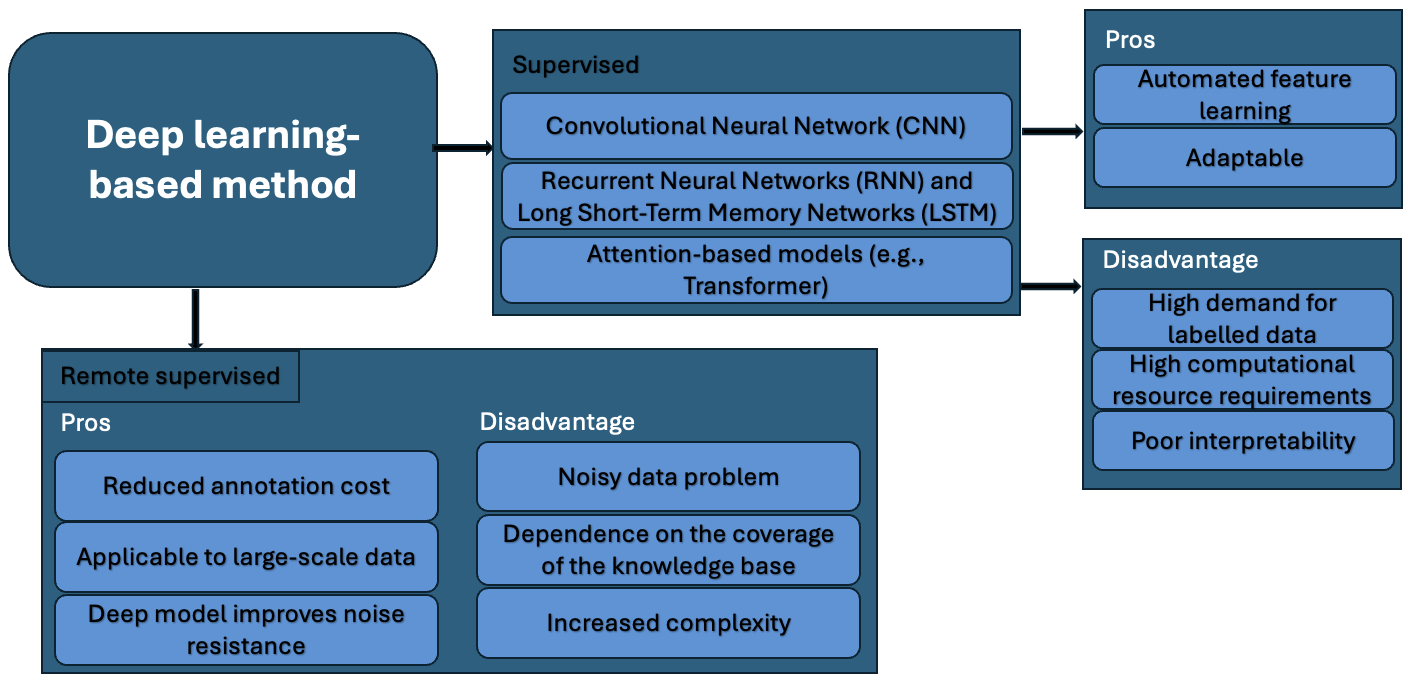}
    \caption{Deep learning-based methods}
    \label{Deep-learning-based-method}
\end{figure}

With the rise of deep learning (DL), the field of NLP has experienced a revolutionary change. 
Deep learning methods, shown in Figure \ref{Deep-learning-based-method}, have greatly improved the performance of relation extraction by automatically learning to extract complex features from unlabelled data through neural network models. Compared with traditional rule-based 
approaches, deep learning is better able to adapt to the \textit{diversity} of languages and performs particularly well when dealing with large-scale data \cite{8416973}.

\textbf{Supervised methods}

In a supervised learning framework, deep learning models use large amounts of \textbf{annotated} data to automatically learn features in text through a complex neural network structure. The following are several commonly used deep learning models and their applications for extraction:

\textbf{Convolutional Neural Network (CNN)}: The application of CNN in relationship extraction is mainly focused on processing sentence-level tasks. Using the sliding window, CNN can effectively extract local contextual information in a sentence, especially in capturing \textit{short-distance dependencies} between words. Because CNN can process data in parallel, they are very suitable for training \textit{large-scale} datasets and therefore have been widely adopted in practical applications \cite{zeng2014relation}.

\textbf{Recurrent Neural Networks (RNNs)}  and \textbf{Long Short-Term Memory Networks (LSTMs)}: RNNs and their variant LSTMs can capture temporal information in sequential data and are especially suitable for dealing with \textit{long}-distance dependencies. LSTM solves the problem of gradient disappearance that occurs in traditional RNNs when dealing with long sequences through the introduction of memory units and the gating mechanism, which enables it to better remember the previous contextual information, thus improving the accuracy of relationship extraction \cite{zhang2015relationclassificationrecurrentneural}.

\textbf{Attention-based models} (e.g., Transformer): The introduction of the attention mechanism further improves the performance of deep learning models. The Transformer model is able to process each word while paying attention to the \textit{contextual} information of all other words in the sentence through the \textit{self-attention} mechanism, which makes the model perform much better in capturing complex relationships  \cite{vaswani2017attention}. Especially when dealing with long sentences and complex contexts, Transformer shows its advantages and has become one of the most popular deep learning models.

\textbf{Advantages:}
\begin{itemize}
\item \textbf{Automated feature learning}: Instead of manually designing features, deep learning models can automatically learn the complex features required for relation extraction from text, which greatly simplifies the process of model development.
\item \textbf{Adaptable}: Deep learning models are able to handle a wide range of complex linguistic phenomena, including long-distance dependencies, nested structures, and non-linear relationships, making them particularly good at diverse natural language processing.
\item \textbf{Highly scalable}: Since deep learning models can be trained in parallel on large-scale datasets, they are highly scalable and suitable for handling large-scale real-world application scenarios.
\end{itemize}

\textbf{Disadvantages:}
\begin{itemize}
\item \textbf{High demand for labelled data}: The training of deep learning models relies on a large amount of labelled data, which may be difficult to obtain in some domains. If the labelled data is insufficient, the model may suffer from overfitting problems, which may affect its generalisation ability.
\item \textbf{High computational resource requirements}: The training of deep learning models usually requires a large amount of computational resources, especially for complex models (e.g., BERT), where training time and resource consumption may become a bottleneck in the application.
\item \textbf{Poor interpretability}: Although deep learning models perform well in terms of performance, their internal working mechanisms are often difficult to explain, which can be a significant drawback for some application scenarios that require a high degree of transparency.
\end{itemize}

\textbf{Remote/Distant supervised methods}

Remote supervised learning is a \textit{weakly supervised} approach to automatically generate annotated data by aligning text with an existing knowledge base. Remote/distance supervision assumes that if there is a known relationship between two entities in the \textit{knowledge base}, 
any sentence containing both entities may express that relationship. While this assumption simplifies the annotation process, it also introduces a large amount of noisy data, as not all sentences containing these entities express the relevant relationship \cite{mintz2009distant,Yuan-etal-2019-distant}.


\textbf{Advantages:}
\begin{itemize}
\item \textbf{Reduced annotation cost}: Remote supervision greatly reduces the cost of manual annotation by automatically generating annotated data using an existing knowledge base, making large-scale relational extraction possible.
\item \textbf{Applicable to large-scale data}: Remote supervision methods can handle large-scale unlabelled corpora, which makes them valuable in constructing large-scale knowledge graphs.
\item \textbf{Deep model improves noise resistance}: By combining deep learning techniques, the remotely supervised relation extraction model can better handle noisy data and improve the accuracy of extraction results.
\end{itemize}

\textbf{Disadvantages:}
\begin{itemize}
\item \textbf{Noisy data problem}: Since the assumptions of remote supervision tend to be too loose, it leads to a large amount of noise in the generated annotated data, which poses a challenge to the training of the model.
\item \textbf{Dependence on the coverage of the knowledge base}: The effectiveness of remote supervision is highly dependent on the coverage and quality of the knowledge base; if the knowledge base is not sufficiently comprehensive or accurate, the extraction results of the model will be limited.
\item \textbf{Increased complexity}: While the introduction of deep learning techniques has improved model performance, it has also increased model complexity, making the training and inference process more time-consuming and labour-intensive.
\end{itemize}

\subsection{Model Card Generation}

In this subsection, we introduce  the very few work available, \textit{to our best knowledge}, on LLM model card generation.

The first work is from \cite{singh2023unlocking} who introduced a dataset to facilitate model training on model card generation. The data set contains 500 QA pairs for 25 learning models. The metadata aims to cover the model model training configurations, datasets used, biases, architecture details. Their investigation on ChatGPT, LLaMa, and Galactica identified the performance gap between these models and humans on factual text responses.

The other two very recent works from this year are from \cite{liu-etal-2024-automatic} and \cite{yang2024report}. 
\cite{liu-etal-2024-automatic} further extended the dataset size to cover 4.8k model cards and 1.4k data cards using their two-step retrieval processing (CardBench and CardGen). 
Instead, the work from \cite{yang2024report} (Report Cards) features on three criteria they designed: specificity, faithfulness, and interpretability. In addition, they proposed a baseline algorithm that does not need human supervision. 

In comparison to these works, we propose to use different methodology by leveraging \textit{dependency parsing} to extract model names and their (applications, licences). 


\section{\textsc{AutoLLM-Card} Methodology} 
\label{method}

\begin{figure}
    \centering
    \includegraphics[width=0.97\linewidth]{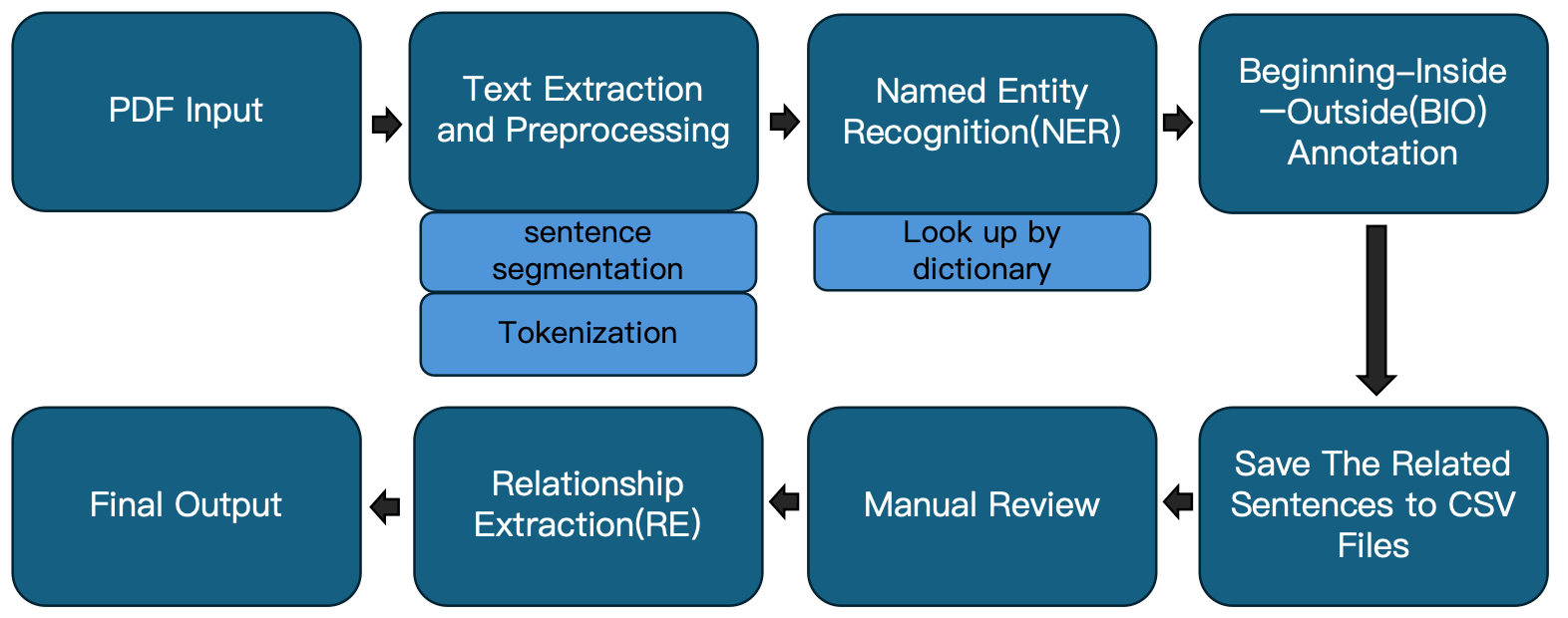}
    \caption{The pipeline of our \textsc{AutoLLM-Card} method}
    \label{pipeline}
\end{figure}
In Figure \ref{pipeline}, the \textbf{names} of large language models and the relationships between their \textbf{applications} or \textbf{licences} are extracted using \textsc{AutoLLM-Card} methodology from academic literature using natural language processing techniques based on the \textit{dependency parsing} approach. 
Dependency parsing is a technique that reveals syntactic dependencies between words in a sentence. It can systematically identify and extract key information describing large language models in the literature and convert it into structured relational data for further analysis.

\begin{figure}
    \centering
    \includegraphics[width=0.99\linewidth]{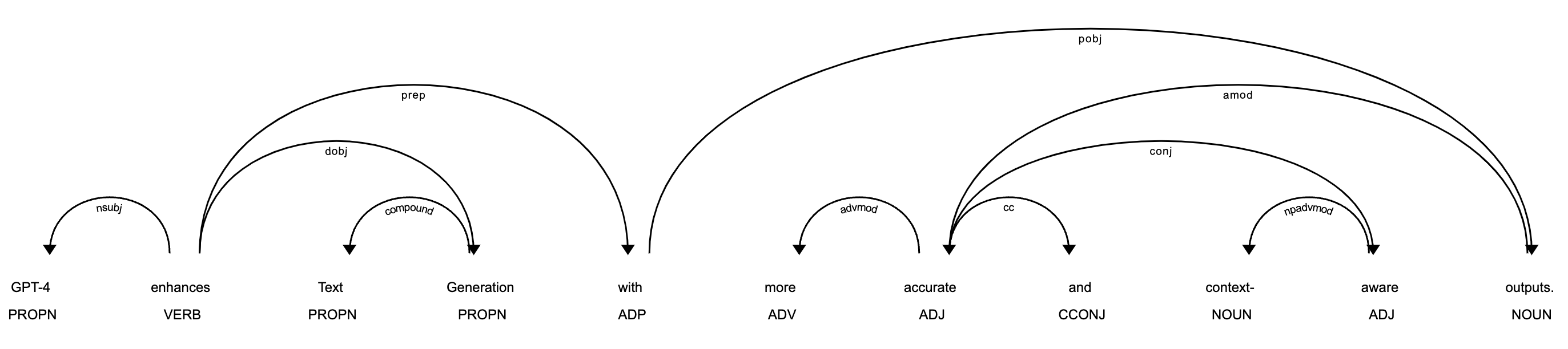}
    \caption{Dependency tree for the active voice processing sentence: \textit{GPT-4 enhances text generation with more accurate and context-aware outputs}.}
    \label{GPT4}
\end{figure}

\begin{figure*}
    \centering
    \includegraphics[width=0.99\linewidth]{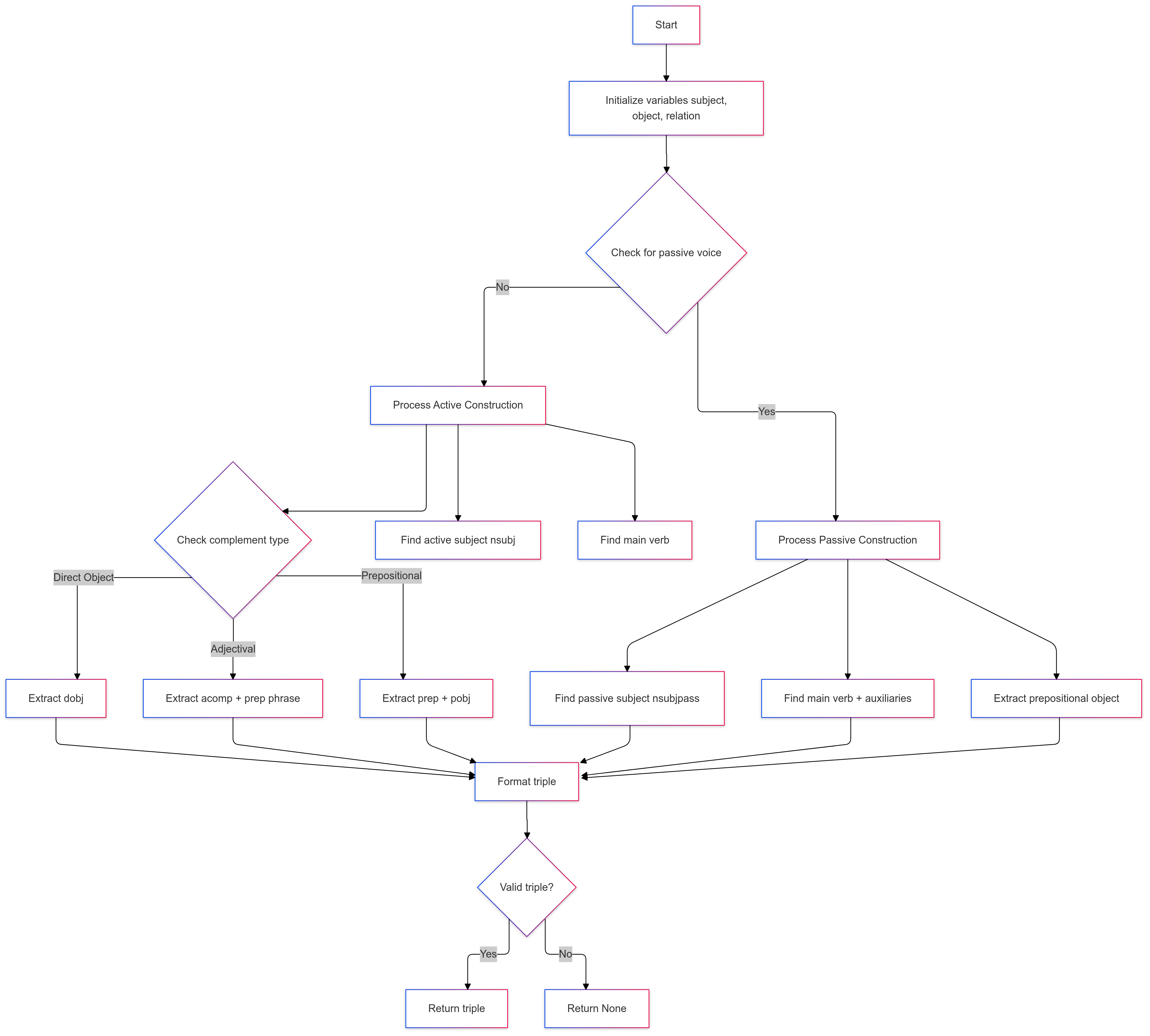}
    \caption{Dependency Tree Based Rules on Passive and Active Voice Syntax.}
    \label{fig:dependency-tree-rules}
\end{figure*}

\begin{figure}
    \centering
    \includegraphics[width=1\linewidth]{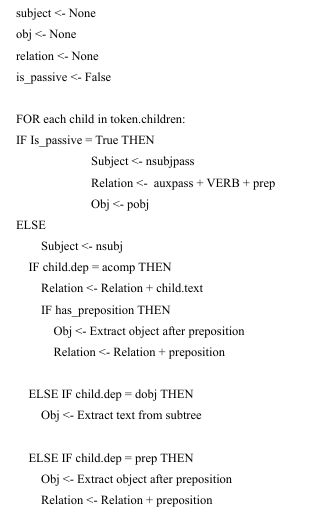}
    \caption{Pseudo Code for Algorithm Illustration}
    \label{fig:PseudoCode}
\end{figure}
\subsection{Data Construction and Relation Extraction}
 
\subsubsection{Data collection and pre-processing}
The first step of the study was to extract text from PDF documents from multiple sources and to ensure comprehensive coverage of the document's content, including body text, footnotes, and graphical descriptions. We removed some unnecessary structural information from the PDF document. 
The extracted text was subjected to \textit{disambiguation} and lexical \textit{annotation} using the \textbf{spaCy} library \cite{honnibal2020spacy} \footnote{\url{https://spacy.io/}}. 
Segmentation breaks down sentences into separate word units (tokens), while lexical annotation assigns each token a corresponding lexical tag (POS tag), such as noun (NOUN), verb (VERB), preposition (ADP), adverb (ADV), etc. SpaCy will use these POS tags to extract dependency relation.
As shown in Figure \ref{BERT}, in the sentence `BERT was released under the Apache licence 2.0', where `BERT' is tagged as a proper noun (PROPN), `was' is tagged as an auxiliary verb (AUX), and `released' is tagged as a verb (VERB). We removed all words ``the'' because we do not think they are useful for the relation extraction task. 
Next, we introduce the relation extraction using dependency trees.

\subsubsection{Dependency Tree Construction using spaCy}
Dependency annotation is the process of parsing the grammatical structure of a sentence and labelling each word in the sentence with its grammatical dependency. These dependency tags (e.g., nsubjpass, auxpass, prep, etc.) can help us understand the main \textit{stem} and \textit{modifiers} in a sentence. 
For example, in the sentence: \textit{BERT was released under the Apache licence 2.0}, where \textit{BERT} is as the subject (nsubjpass) depends on the verb \textit{released}.

After the annotation of dependency relations, we can represent the syntactic structure of a sentence as a \textbf{dependency tree}. 
The root node of a dependency tree is usually the predicate verb of a sentence. 
By constructing a dependency tree, we can visualise the grammatical structure and logical relationships of a sentence. For example, in the sentence: \textit{BERT was released under the Apache licence 2.0}, where \textit{released} is the root node and other words depend on it.

The overall diagram on how we applied dependence parsing rules is presented in Figure \ref{fig:dependency-tree-rules}. A step-by-step interpretation is as follows.

First, we set four initial parameters, subject, object(obj), relation, and is\_passive. 
Through these parameters, we respectively stored the sentence subject information into the subject, the direct object and object part into the obj, the root part between subject and obj as relation, and identified whether the sentence is in passive voice. Secondly, if we identify the sentence as passive voice, we process the passive construction in terms of finding the passive subject, main vern with its auxillaries part, and prepositional object to form the relational triple. 
If we identify the sentence as the active construction we find the subject part first and check the active complement type in terms of direct object, adjectival, and prepositional respectively to extract direct object(dobj), adjective complement(acomp) with prep phrase, prep phrase with prep object(pobj) to form the triple. 
Finally, we confirm that the triples are valid.
The algorithm illustration using pseudo-code is in Figure \ref{fig:PseudoCode}.


To give interpretation examples, in the left hand of the diagram in Figure \ref{fig:dependency-tree-rules}, for ``Check complement type'':
\begin{itemize}
    \item 1. Direct object => e.g. ``ChatGPT improves NER’’ => dobj(improves, NER)
    \item 2. Adjectival: e.g. ``ChatGPT is good at NER’’ => acomp(is, good); prep phrase: at NER
    \item 3. Prepositional: e.g. ``ChatGPT applies at NER task’’ => prep+pobj: at + pobj(NER task)
\end{itemize}

For more examples, ``dobj’’ is the relation (enhances, Text Generation) in Figure \ref{GPT4}; 
``acomp’’ is the relation  (is, effective), ``prep’’ is the relation (effective, in), and ``pobj’’ is the relation (in, Named Entity Recognition)  in Figure \ref{ERNIE}.

\subsection{Rules Applied for Specific Relations}
As a \textit{pilot} study, we first applied rules on LLMs and their (applications, licences).
To extract the relationship between LLM \textbf{names} and \textbf{application} scenarios or \textbf{licences} from sentences, a set of \textbf{rules} was developed. These rules identify and extract the relationships between model names, application scenarios, and licences based on the \textit{syntactic} structure of the sentence.
%
Dependency analysis can not only reveal subject-object relationships but also capture \textit{modification} relationships and compound structures. 
As shown in Figure \ref{GPT4}, the sentence: \textit{GPT-4 enhances Text Generation with more accurate and context-aware outputs}, where \textit{enhances} as a core verb forms the main semantic relationship with the subject \textit{GPT-4} and the object \textit{Text Generation}, while \textit{enhances} forms the main semantic relationship with \textit{GPT-4} and the object \textit{Text Generation}. 
Through these parses, we are able to systematically extract and structure this information, providing a basis for more advanced tasks such as information extraction and relational modelling.
In the next section, we introduce our experimental settings in detail and the outcomes.

\section{Experiment} 
\label{Experiment}
The implementation process of this work is divided into the following \textbf{key steps}: multi-source data extraction, keyword identification and context capture, data screening and organisation, entity recognition annotation (NER), dependency resolution and relationship extraction, and visualisation and querying of relationship triples. 
The following is a detailed description of each step and its logical relationship in the overall process.

\subsection{Settings}

Complex data formats require complex logic to parse and understand, the Portable Document Format (PDF) is one of the most demanding formats because it is both a data exchange format and a presentation format and has a particularly rigorous tradition of supporting interoperability and consistent presentation \cite{anantharaman2023polydoc}. 
This work mainly uses the  \textbf{PyMuPDF} python library to extract the text of PDF. 
106 academic papers were processed by defining three dictionaries- LLMs name, licence, and application.
11,051 sentences were extracted through dictionary lookup, and the dataset was constructed through manual review of the final selection of 129 sentences with a link between the name and the licence, and 106 sentences with a link between the model name and the application.

In order to cope with document contents of different \textit{formats} and \textit{complexity}, a multi-source data extraction strategy is employed. 
First, we extract text content directly from PDF files using an efficient text parsing tool, which is particularly good at handling documents containing complex structures, such as footnotes and figure illustrations.
If the parsing process encounters difficulties or the page cannot be parsed, the system will automatically switch to an alternate text extraction tool. 
This \textbf{multi-level extraction} method ensures the completeness and accuracy of the document data and lays a solid foundation for subsequent analyses.

After the text extraction was completed, an automated \textbf{keyword search} tool was developed to accurately identify LLM names, licence types and model application scenarios mentioned in the literature. The core functions of the tool include \textbf{keyword} matching and \textbf{contextual} analysis. Firstly, the tool scans the extracted text with a \textit{predefined} list of keywords (e.g. specific LLM names, licence types, etc.) and marks all matching keywords. 
Next, by analysing the \textit{dependencies} and \textit{grammatical} structure of the sentences in which the keywords are located, the tool is able to automatically extract contextual information related to the keywords, which includes the content of several sentences before and after, ensuring that the actual application scenarios of the keywords are captured, rather than just a single term match. 
Through this approach, the system is able to identify sentences containing key information in the literature and correlate these sentences with their contexts, providing reliable basic data for subsequent data screening, relationship extraction and structured data construction. This process significantly improves the accuracy and comprehensiveness of information extraction.

When the automated keyword recognition was completed, the extracted sentences were \textit{manually screened}. Although the automated tool performed well in the initial screening, manual intervention helped to further improve the accuracy and relevance of the data. 
The manual review enabled the identification and correction of misclassifications or false matches that may have occurred during the automation process. This step lays the foundation for structuring and standardising the data and directly affects the accuracy of the subsequent entity identification annotation.

\begin{table}[h!]
\centering
\begin{tabularx}{0.49\textwidth}{|l|c|l|}
\hline
Subject&  Relation&Object
\\ \hline
GPT-3&  is used for&Text Generation
\\ \hline
GPT-3&  powers&Conversational Agents
\\ \hline
GPT-4&  enhances&Text Generation
\\ \hline
GPT-4&  improves&Question Answering
\\ \hline
 GPT-4& is used in&Creative Writing
\\\hline
 BERT& is essential for&Text Classification
\\\hline
 BERT& is applied in&Named Entity Recognition ( NER )
\\\hline
 RoBERTa& improves&Text Classification
\\\hline
 RoBERTa& is used in&Knowledge Graph Construction
\\\hline
 ...& &\\\hline
\end{tabularx}
\caption{Example of Relational extraction datasets for large language model and application}
\label{application table}
\end{table}

\begin{table}[h!]
\centering
\begin{tabularx}{0.49\textwidth}{|l|X|X|}
\hline
Subject&  Relation&Object
\\ \hline
GPT-4&  is licenced under&a proprietary agreement by OpenAI
\\ \hline
BERT&  was released under&Apache licence 2.0
\\ \hline
RoBERTa&  is available under&Apache licence 2.0
\\ \hline
XLNet&  is distributed under&Apache licence 2.0
\\ \hline
 T5& is subject to&Apache licence 2.0
\\\hline
 DistilBERT& was released with&Apache licence 2.0
\\\hline
 Turing - NLG& is licenced under&a proprietary agreement with Microsoft
\\\hline
 Transformer - XL& was released under&Apache licence 2.0
\\\hline
 OpenAI Codex& is distributed with&a proprietary licence from OpenAI\\\hline
 ...& &\\\hline
\end{tabularx}
\caption{Example of Relational extraction dataset for large language model and licence}
\label{licence table}
\end{table}


After multi-source data extraction, keyword recognition, context capture, manual screening, NER annotation, and dependency parsing, the following \textbf{two datasets} are successfully constructed,
shown in Table \ref{application table} and \ref{licence table}  with examples: \textbf{Application} dataset of big language models: this dataset contains the performance of multiple big language models in different application scenarios, and combines with dependency parsing techniques to reveal the application of each model. \textbf{Licensing} dataset of big language models: this dataset records the types of licences adopted by each model and its legal framework. It clarifies the correspondence between models and licences by parsing the associated sentences.

\subsection{Rule Formulation}

This work performed dependency parsing on sentences using the \textbf{spaCy} library to extract relationships between model names and application scenarios or licences. Dependency parsing reveals the syntactic relationships between words in a sentence, allowing us to extract explicit relational triples (e.g. subject-predicate-object) from complex syntactic structures. By developing a set of rules for dealing with \textbf{active} and \textbf{passive} \textbf{voice}, we can systematically extract relevant relations from sentences with different syntactic structures and transform them into structured data. The generated \textbf{relation triples} provide the core data for subsequent \textit{visualisation} and \textit{query} functions.

\subsubsection{Active voice processing}

\begin{figure}
    \centering
    \includegraphics[width=0.99\linewidth]{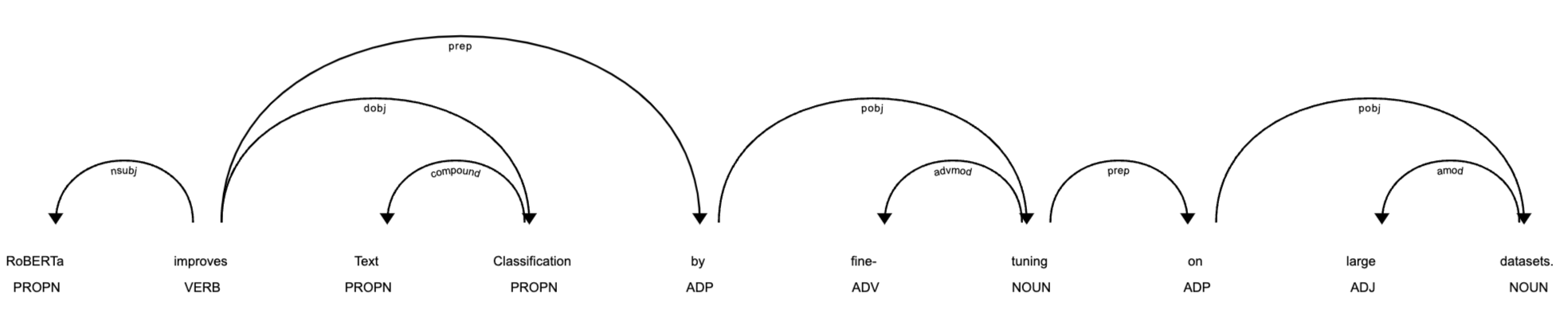}
    \caption{Dependency tree for the active voice processing sentence: \textit{RoBERTa improves text classification by fine-tuning on large datasets}.}
    \label{RoBert}
\end{figure}

\begin{figure}
    \centering
    \includegraphics[width=0.99\linewidth]{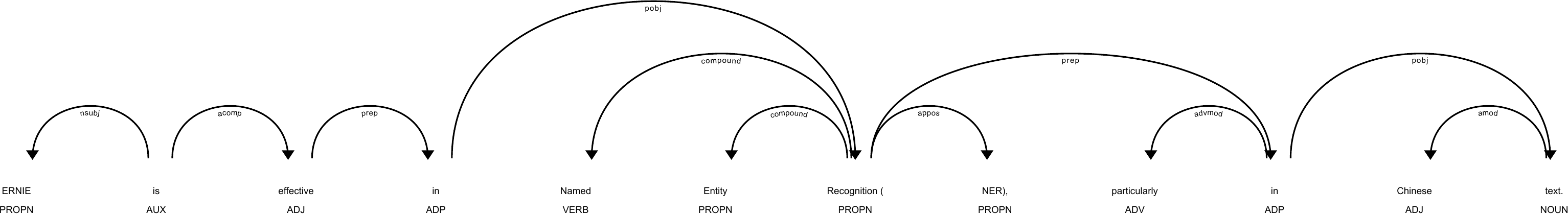}
    \caption{Dependency tree for the active voice processing sentence: \textit{ERNIE is effective in Named Entity Recognition (NER), particularly in Chinese text}.}
    \label{ERNIE}
\end{figure}

Identifying Subjects: for active voice sentences, the sentence's subject is usually the name of a large language model. Dependency parsing enables the identification of the entity that is the subject of the sentence and identifies it as the subject part of the relation. 
Identifying predicates: the system identifies the predicate verbs of the sentence, which typically describe the behaviour or application of the model. For example, the verb may indicate a release, application, or integration of the model. At this point, the system captures both auxiliary verbs and modifiers associated with the predicate verb to ensure the integrity of the predicate. 
Identifying \textit{objects} or \textit{prepositional} phrases: the system looks for direct objects or prepositional phrases associated with the predicate verb. These components typically describe the application scenario or licence type of the model. By identifying these constituents, the system can associate a model name with its application or licence.
As shown in Figure \ref{RoBert}, in the sentence \textit{RoBERTa improves text classification by fine-tuning on large datasets}, the system recognises \textit{RoBERTa} as the subject, \textit{improves} as the predicate verb, and \textit{text classification} as the direct object, which further specifies the application of the model RoBERTa in text classification.
In Figure \ref{ERNIE}, the root is \textit{effective}, which is an adjective. \textit{ERNIE} is the subject, \textit{is} is the auxiliary verb, and \textit{in Named Entity Recognition (NER)} is a prepositional phrase describing where ERNIE is effective. The phrase \textit{particularly in Chinese text} further modifies the effectiveness of ERNIE.

\subsubsection{Passive voice processing}

Identifying passive subjects: For passive sentences, the system first detects passive auxiliary verbs (e.g., \textit{was}) and identifies the passive subject, which is usually the name of the large language model.

\begin{figure}[th]
    \centering
    \includegraphics[width=0.99\linewidth]{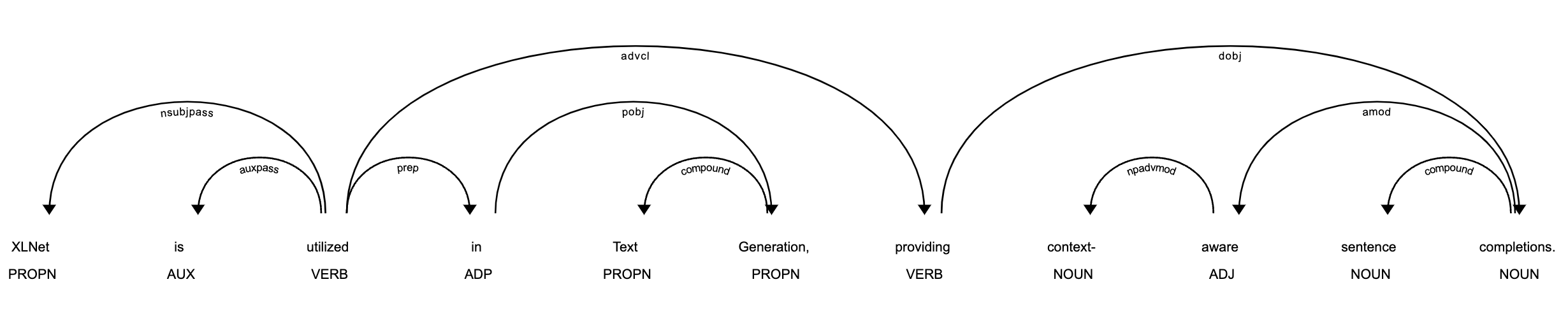}
    \caption{Dependency tree for the passive voice processing sentence \textit{XLNet is utilised in Text Generation, providing context-aware sentence completions}.}
    \label{XLNet}
\end{figure}

Identifying \textit{predicates} and \textit{prepositional} phrases: In passive voice, predicate verbs are usually combined with prepositional phrases to describe the association of the model with a licence or application scenario. The system identifies these prepositional phrases and their objects through dependency parsing to construct a complete relationship.
In Figure \ref{XLNet}, the root is \textit{utilised}, which is the main verb. \textit{XLNet} is the subject, \textit{is} is the auxiliary verb, and \textit{in Text Generation} is a prepositional phrase indicating where XLNet is utilised. The phrase \textit{providing context-aware sentence completions} acts as a participial phrase explaining what XLNet does when utilised.

\begin{figure}[th]
    \centering
    \includegraphics[width=0.99\linewidth]{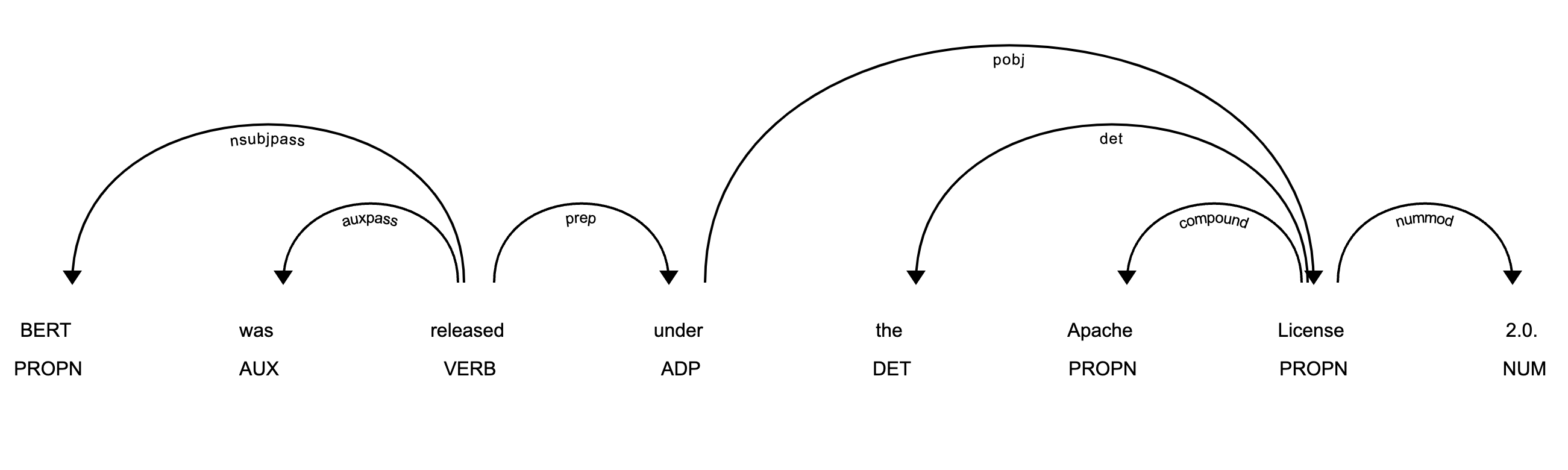}
    \caption{Dependency tree for the passive voice processing sentence: \textit{BERT was released under the Apache licence 2.0.}}
    \label{BERT}
\end{figure}

In Figure \ref{BERT}, the root is \textit{released}, which is the main verb. \textit{BERT} is the subject, \textit{was} is the auxiliary verb, and \textit{under the Apache licence 2.0} is a prepositional phrase indicating the condition under which BERT was released.

With these rules, we were able to systematically extract clear \textit{relational} \textit{triples} (subject-predicate-object) from complex sentence structures that accurately reflect the associations between model names and their application scenarios or licences. The development and implementation of these rules ensured that the system was able to extract relevant information flexibly and accurately when confronted with sentences with different syntactic structures.

\subsubsection{Evaluation}

In practice, this relationship extraction method based on dependency parsing demonstrated excellent performance in the dataset we constructed. 
In order to comprehensively evaluate the effectiveness of the method, we used several commonly used evaluation metrics, including Accuracy, Recall, Precision, and F1-Score. These metrics are crucial in measuring the overall performance of a relational extraction model, and can comprehensively reflect the performance of the model in different dimensions, thus providing a scientific basis for model optimisation and improvement. The evaluation results on the test dataset are as shown in Table \ref{table:metrics}.
The evaluation results indicate that the method has a high degree of accuracy and consistency in identifying and extracting relationships related to LLMs, and especially excels in dealing with complex syntactic structures. 

\begin{table}[h!]
\centering
\begin{tabular}{|l|c|}
\hline
\textbf{Metric}     & \textbf{Value} \\ \hline
Accuracy            & 0.92           \\ \hline
Recall              & 0.88           \\ \hline
Precision           & 0.90           \\ \hline
F1 value (F1-Score) & 0.89           \\ \hline
\end{tabular}
\caption{{Evaluation metrics for the test dataset}
}
\label{table:metrics}
\end{table}

\subsection{Visualisation}

\begin{figure}[th]
    \centering
    \includegraphics[width=0.99\linewidth]{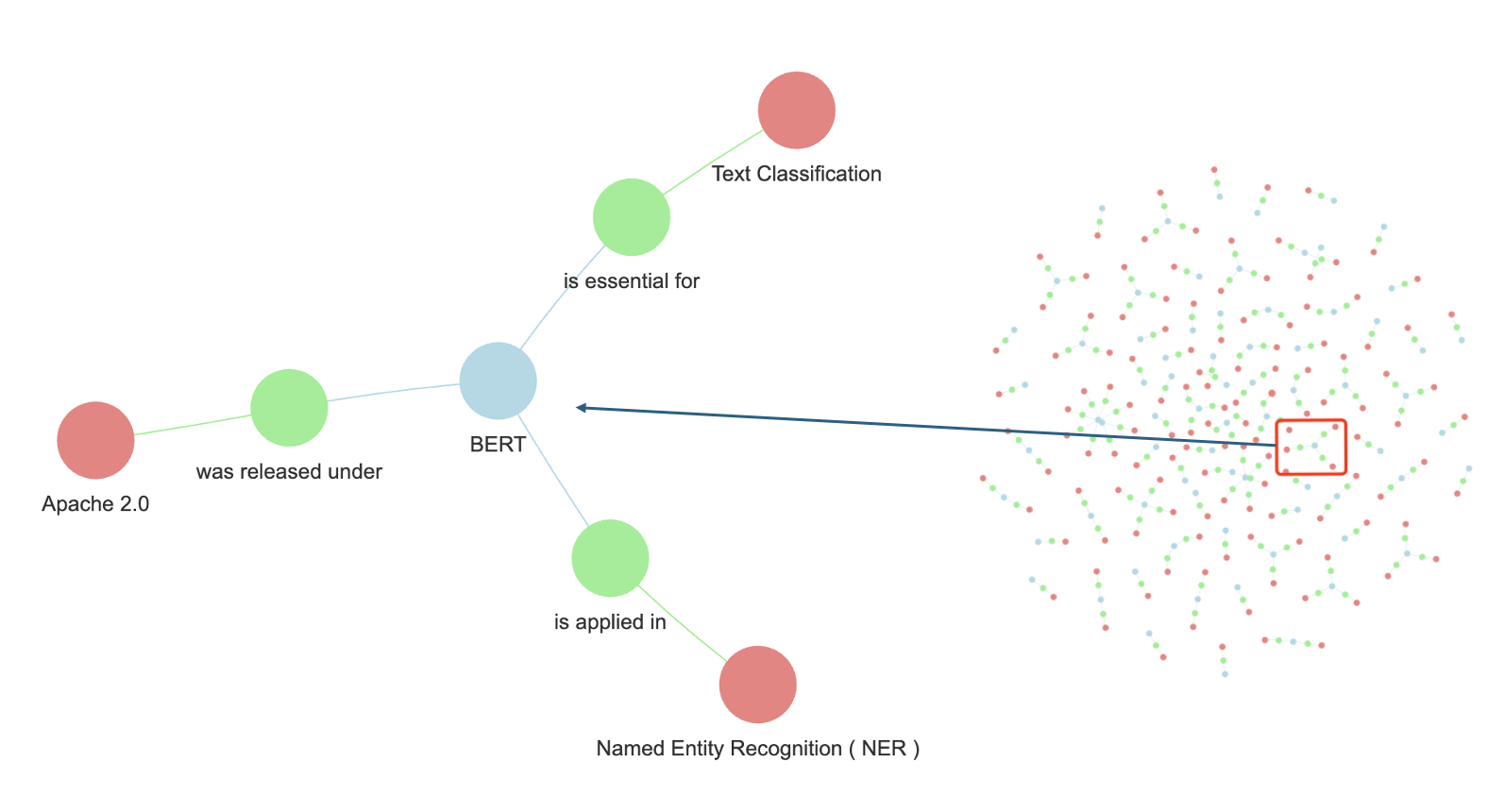}
    \caption{A visual summary of various large language models, applications, and licences. \textbf{Blue} represents the name of the model, \textbf{green} represents the relationship between the model name and the licence and application entities, and \textbf{red} represents the application and licence of the large language model.}
    \label{fig:KG_visual_example}
\end{figure}
Figure \ref{fig:KG_visual_example} illustrates the output \textbf{visualisation} of extracted relations where users can drag the extracted components.  
By developing and applying this series of automated data processing tools, we have successfully constructed a structured \textbf{dataset} containing LLM names, licence types, and application scenarios. This dataset not only provides a solid foundation for subsequent relationship extraction and knowledge graph construction but also demonstrates the power of modern automated text processing techniques in academic research.

From the evaluation results, the model performs well in all indicators, especially in identifying relationships in complex syntactic structures. This excellent performance not only verifies the effectiveness of the relationship extraction method based on dependency parsing but also highlights the broad applicability of the method in dealing with diverse literature data.

Following the completion of the extraction of the relation triples, we further demonstrate these relations by constructing a \textbf{graphical} database. Specifically, we use the NetworkX and PyVis libraries to visualise the extracted relation triples as graph structures and allow the graphs to be queried via a Depth-First Search (DFS) algorithm. This step not only transforms complex textual relationships into intuitive graphical presentations but also provides flexible querying functionality that allows researchers to easily retrieve and explore the association information of specific models under different application scenarios and licences.


\section{Conclusions and Future Work}
\label{Conclusions and Future Work}
\subsection{Conclusions}

As a pilot study, the main goal of this work is to validate the basic feasibility of the proposed method \textsc{AutoLLM-Card}, i.e., automatically extract LLMs in published literature and their model cards of model details and their affiliated information. 
We have successfully constructed \textbf{two core datasets} using dependency parsing and relation extraction techniques: one on the \textit{application} scenarios of big language models, and the other on the types of \textit{licences} used in these models. 
These datasets provide a solid foundation for further research. 

In order to further validate the utility of the method, we also conducted real-world testing of the extracted relations and entities using locally deployed large language models. 
These big language models performed well in handling natural language tasks, especially in the absence of specialised training data, and were able to accurately identify and predict many complex linguistic relations by virtue of their powerful zero-shot learning capabilities. However, despite the model's good performance in initial tests, there is still room for improvement in its predictions. In order to improve the model's performance in various complex situations, further \textit{instruction training} is essential (e.g. methods from \cite{cui-etal-2023-medtem2,ren2024synthetic4healthgeneratingannotatedsynthetic,micheletti2024explorationmaskedcausallanguage,belkadi2024generatingsyntheticfreetextmedical}). Especially when it comes to literature in specialised areas, the model needs targeted training to enhance its understanding of specific terms and complex grammatical structures, thus ensuring the accuracy and applicability of the prediction results.

 \subsection{Future work}

Prospectively, an important direction for this research lies in utilising the experimental results of this work, i.e., \textit{fine-tuning} the training of the new model with the relational and entity datasets extracted from this study. 
These datasets were accurately labelled and processed to provide ideal training material for further optimisation of the model. 
Future research can design more \textit{complex} and \textit{targeted} training tasks to enable the big language model to better grasp \textit{domain}-specific linguistic features and enhance its performance in real-world applications. 
In addition, as the amount of data increases, we can explore the use of these relational and entity data to construct more complex knowledge graphs to further enhance the model's \textit{reasoning} and knowledge management capabilities. 
Such efforts not only help to enhance the understanding and generative capabilities of the models, but also provide possibilities for a wider range of application scenarios in the future.

We plan to carry more model comparisons with the very recent works such as \cite{singh2023unlocking,liu-etal-2024-automatic,yang2024report}. We are also implementing new methods regarding our NER component of the \textsc{AutoLLM-Card} pipeline.

\section*{Limitations}
As proof of concept investigation, the scope of validation in this study is limited, focusing on a small number of literature and specific tasks, and has not yet been fully tested on larger datasets. 
Although the preliminary results show the effectiveness of the method \textsc{AutoLLM-Card}, future research needs to expand the \textit{size} of the validation dataset and apply more stringent evaluation criteria to comprehensively examine the \textit{robustness} and applicability of the method.

\section*{Acknowledgement}
This updated version would not have been possible without the scientific and critical review and feedback from Prof Suzan Verbene. 

\appendix

\section{Appendix}
\begin{figure}[th]
    \centering
    \includegraphics[width=0.99\linewidth]{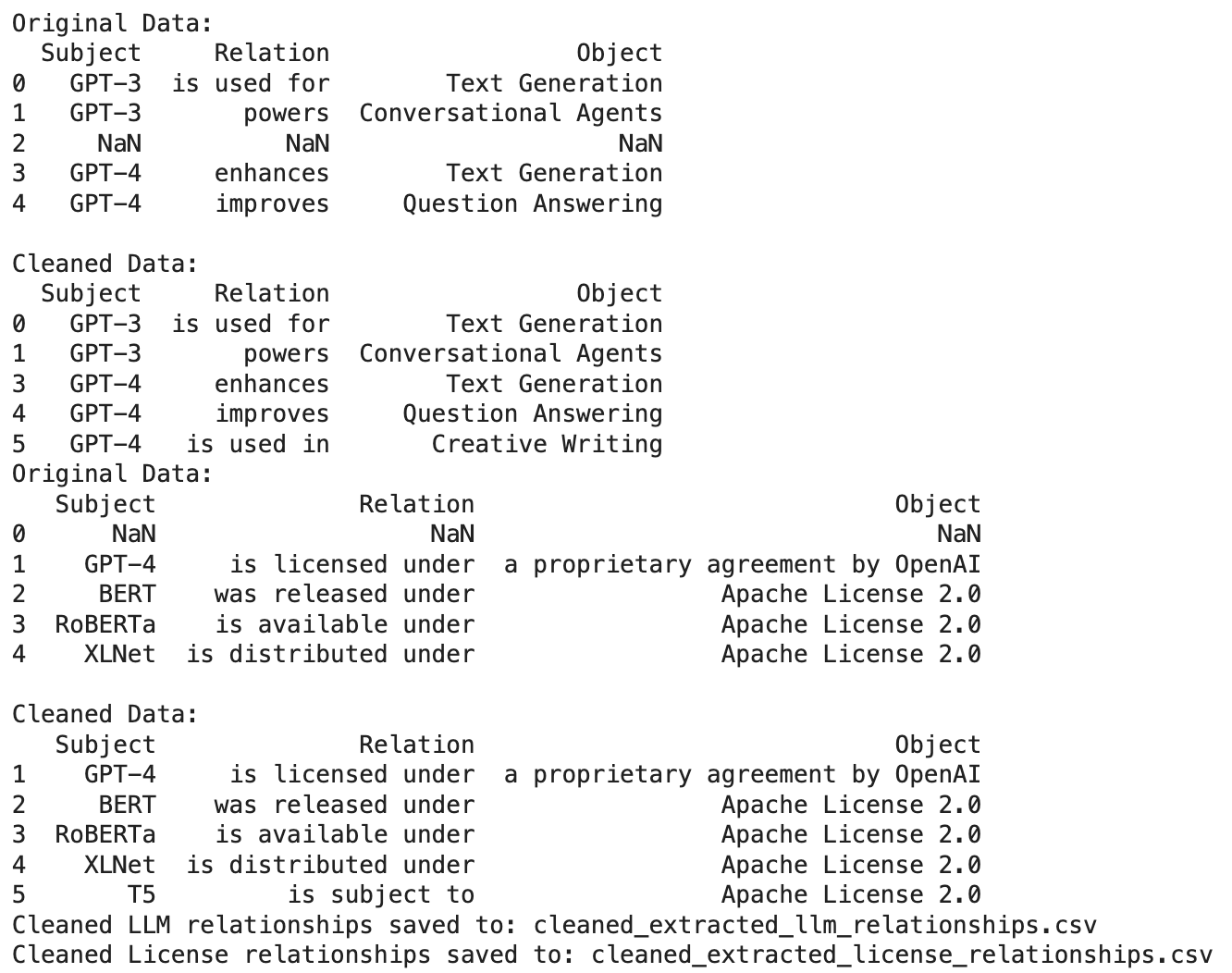}
    \caption{data cleaning example after filtering out the non-relation entries}
    \label{data-cleaning}
\end{figure}

\bibliographystyle{ACM-Reference-Format}
\bibliography{sample-bibliography,nodalida2023,references}


\begin{thebibliography}{28}


\ifx \showCODEN    \undefined \def \showCODEN     #1{\unskip}     \fi
\ifx \showDOI      \undefined \def \showDOI       #1{#1}\fi
\ifx \showISBNx    \undefined \def \showISBNx     #1{\unskip}     \fi
\ifx \showISBNxiii \undefined \def \showISBNxiii  #1{\unskip}     \fi
\ifx \showISSN     \undefined \def \showISSN      #1{\unskip}     \fi
\ifx \showLCCN     \undefined \def \showLCCN      #1{\unskip}     \fi
\ifx \shownote     \undefined \def \shownote      #1{#1}          \fi
\ifx \showarticletitle \undefined \def \showarticletitle #1{#1}   \fi
\ifx \showURL      \undefined \def \showURL       {\relax}        \fi
\providecommand\bibfield[2]{#2}
\providecommand\bibinfo[2]{#2}
\providecommand\natexlab[1]{#1}
\providecommand\showeprint[2][]{arXiv:#2}

\bibitem[Anantharaman et~al\mbox{.}(2023)]%
        {anantharaman2023polydoc}
\bibfield{author}{\bibinfo{person}{Prashant Anantharaman}, \bibinfo{person}{Robert Lathrop}, \bibinfo{person}{Rebecca Shapiro}, {and} \bibinfo{person}{Michael~E Locasto}.} \bibinfo{year}{2023}\natexlab{}.
\newblock \showarticletitle{PolyDoc: Surveying PDF Files from the PolySwarm network}. In \bibinfo{booktitle}{\emph{2023 IEEE Security and Privacy Workshops (SPW)}}. IEEE, \bibinfo{pages}{117--134}.
\newblock


\bibitem[Belkadi et~al\mbox{.}(2024)]%
        {belkadi2024generatingsyntheticfreetextmedical}
\bibfield{author}{\bibinfo{person}{Samuel Belkadi}, \bibinfo{person}{Libo Ren}, \bibinfo{person}{Nicolo Micheletti}, \bibinfo{person}{Lifeng Han}, {and} \bibinfo{person}{Goran Nenadic}.} \bibinfo{year}{2024}\natexlab{}.
\newblock \bibinfo{title}{Generating Synthetic Free-text Medical Records with Low Re-identification Risk using Masked Language Modeling}.
\newblock
\newblock
\showeprint[arxiv]{2409.09831}~[cs.CL]
\urldef\tempurl%
\url{https://arxiv.org/abs/2409.09831}
\showURL{%
\tempurl}


\bibitem[Brown(2020)]%
        {brown2020language}
\bibfield{author}{\bibinfo{person}{Tom~B Brown}.} \bibinfo{year}{2020}\natexlab{}.
\newblock \showarticletitle{Language models are few-shot learners}.
\newblock \bibinfo{journal}{\emph{arXiv preprint arXiv:2005.14165}} (\bibinfo{year}{2020}).
\newblock


\bibitem[Cui et~al\mbox{.}(2023)]%
        {cui-etal-2023-medtem2}
\bibfield{author}{\bibinfo{person}{Yang Cui}, \bibinfo{person}{Lifeng Han}, {and} \bibinfo{person}{Goran Nenadic}.} \bibinfo{year}{2023}\natexlab{}.
\newblock \showarticletitle{{M}ed{T}em2.0: Prompt-based Temporal Classification of Treatment Events from Discharge Summaries}. In \bibinfo{booktitle}{\emph{Proceedings of the 61st Annual Meeting of the Association for Computational Linguistics (Volume 4: Student Research Workshop)}}. \bibinfo{publisher}{Association for Computational Linguistics}, \bibinfo{address}{Toronto, Canada}, \bibinfo{pages}{160--183}.
\newblock
\urldef\tempurl%
\url{https://aclanthology.org/2023.acl-srw.27}
\showURL{%
\tempurl}


\bibitem[Devlin et~al\mbox{.}(2019)]%
        {devlin-etal-2019-bert}
\bibfield{author}{\bibinfo{person}{Jacob Devlin}, \bibinfo{person}{Ming-Wei Chang}, \bibinfo{person}{Kenton Lee}, {and} \bibinfo{person}{Kristina Toutanova}.} \bibinfo{year}{2019}\natexlab{}.
\newblock \showarticletitle{{BERT}: Pre-training of Deep Bidirectional Transformers for Language Understanding}. In \bibinfo{booktitle}{\emph{Proceedings of the 2019 Conference of the North {A}merican Chapter of the Association for Computational Linguistics: Human Language Technologies, Volume 1 (Long and Short Papers)}}. \bibinfo{publisher}{Association for Computational Linguistics}, \bibinfo{address}{Minneapolis, Minnesota}, \bibinfo{pages}{4171--4186}.
\newblock
\urldef\tempurl%
\url{https://doi.org/10.18653/v1/N19-1423}
\showDOI{\tempurl}


\bibitem[Honnibal et~al\mbox{.}(2020)]%
        {honnibal2020spacy}
\bibfield{author}{\bibinfo{person}{Matthew Honnibal}, \bibinfo{person}{Ines Montani}, \bibinfo{person}{Sofie Van~Landeghem}, {and} \bibinfo{person}{Adriane Boyd}.} \bibinfo{year}{2020}\natexlab{}.
\newblock \showarticletitle{{spaCy: Industrial-strength Natural Language Processing in Python}}.
\newblock  (\bibinfo{year}{2020}).
\newblock
\urldef\tempurl%
\url{https://doi.org/10.5281/zenodo.1212303}
\showDOI{\tempurl}


\bibitem[Jurafsky(2000)]%
        {jurafsky2000speech}
\bibfield{author}{\bibinfo{person}{Daniel Jurafsky}.} \bibinfo{year}{2000}\natexlab{}.
\newblock \bibinfo{title}{Speech and language processing}.
\newblock
\newblock


\bibitem[Liu et~al\mbox{.}(2024)]%
        {liu-etal-2024-automatic}
\bibfield{author}{\bibinfo{person}{Jiarui Liu}, \bibinfo{person}{Wenkai Li}, \bibinfo{person}{Zhijing Jin}, {and} \bibinfo{person}{Mona Diab}.} \bibinfo{year}{2024}\natexlab{}.
\newblock \showarticletitle{Automatic Generation of Model and Data Cards: A Step Towards Responsible {AI}}. In \bibinfo{booktitle}{\emph{Proceedings of the 2024 Conference of the North American Chapter of the Association for Computational Linguistics: Human Language Technologies (Volume 1: Long Papers)}}, \bibfield{editor}{\bibinfo{person}{Kevin Duh}, \bibinfo{person}{Helena Gomez}, {and} \bibinfo{person}{Steven Bethard}} (Eds.). \bibinfo{publisher}{Association for Computational Linguistics}, \bibinfo{address}{Mexico City, Mexico}, \bibinfo{pages}{1975--1997}.
\newblock
\urldef\tempurl%
\url{https://doi.org/10.18653/v1/2024.naacl-long.110}
\showDOI{\tempurl}


\bibitem[Liu et~al\mbox{.}(2020)]%
        {liu2020roberta}
\bibfield{author}{\bibinfo{person}{Yinhan Liu}, \bibinfo{person}{Myle Ott}, \bibinfo{person}{Naman Goyal}, \bibinfo{person}{Jingfei Du}, \bibinfo{person}{Mandar Joshi}, \bibinfo{person}{Danqi Chen}, \bibinfo{person}{Omer Levy}, \bibinfo{person}{Mike Lewis}, \bibinfo{person}{Luke Zettlemoyer}, {and} \bibinfo{person}{Veselin Stoyanov}.} \bibinfo{year}{2020}\natexlab{}.
\newblock \bibinfo{title}{Ro{\{}BERT{\}}a: A Robustly Optimized {\{}BERT{\}} Pretraining Approach}.
\newblock
\newblock
\urldef\tempurl%
\url{https://openreview.net/forum?id=SyxS0T4tvS}
\showURL{%
\tempurl}


\bibitem[Micheletti et~al\mbox{.}(2024)]%
        {micheletti2024explorationmaskedcausallanguage}
\bibfield{author}{\bibinfo{person}{Nicolo Micheletti}, \bibinfo{person}{Samuel Belkadi}, \bibinfo{person}{Lifeng Han}, {and} \bibinfo{person}{Goran Nenadic}.} \bibinfo{year}{2024}\natexlab{}.
\newblock \bibinfo{title}{Exploration of Masked and Causal Language Modelling for Text Generation}.
\newblock
\newblock
\showeprint[arxiv]{2405.12630}~[cs.CL]
\urldef\tempurl%
\url{https://arxiv.org/abs/2405.12630}
\showURL{%
\tempurl}


\bibitem[Mikolov(2013)]%
        {mikolov2013efficient}
\bibfield{author}{\bibinfo{person}{Tomas Mikolov}.} \bibinfo{year}{2013}\natexlab{}.
\newblock \showarticletitle{Efficient estimation of word representations in vector space}.
\newblock \bibinfo{journal}{\emph{arXiv preprint arXiv:1301.3781}} (\bibinfo{year}{2013}).
\newblock


\bibitem[Mintz et~al\mbox{.}(2009)]%
        {mintz2009distant}
\bibfield{author}{\bibinfo{person}{Mike Mintz}, \bibinfo{person}{Steven Bills}, \bibinfo{person}{Rion Snow}, {and} \bibinfo{person}{Dan Jurafsky}.} \bibinfo{year}{2009}\natexlab{}.
\newblock \showarticletitle{Distant supervision for relation extraction without labeled data}. In \bibinfo{booktitle}{\emph{Proceedings of the Joint Conference of the 47th Annual Meeting of the ACL and the 4th International Joint Conference on Natural Language Processing of the AFNLP}}. \bibinfo{pages}{1003--1011}.
\newblock


\bibitem[OpenAI(2023)]%
        {openai2023gpt}
\bibfield{author}{\bibinfo{person}{R OpenAI}.} \bibinfo{year}{2023}\natexlab{}.
\newblock \showarticletitle{Gpt-4 technical report. arxiv 2303.08774}.
\newblock \bibinfo{journal}{\emph{View in Article}} \bibinfo{volume}{2}, \bibinfo{number}{5} (\bibinfo{year}{2023}).
\newblock


\bibitem[Raffel et~al\mbox{.}(2020)]%
        {raffel2020exploring}
\bibfield{author}{\bibinfo{person}{Colin Raffel}, \bibinfo{person}{Noam Shazeer}, \bibinfo{person}{Adam Roberts}, \bibinfo{person}{Katherine Lee}, \bibinfo{person}{Sharan Narang}, \bibinfo{person}{Michael Matena}, \bibinfo{person}{Yanqi Zhou}, \bibinfo{person}{Wei Li}, {and} \bibinfo{person}{Peter~J Liu}.} \bibinfo{year}{2020}\natexlab{}.
\newblock \showarticletitle{Exploring the limits of transfer learning with a unified text-to-text transformer}.
\newblock \bibinfo{journal}{\emph{Journal of machine learning research}} \bibinfo{volume}{21}, \bibinfo{number}{140} (\bibinfo{year}{2020}), \bibinfo{pages}{1--67}.
\newblock


\bibitem[Ren et~al\mbox{.}(2024)]%
        {ren2024synthetic4healthgeneratingannotatedsynthetic}
\bibfield{author}{\bibinfo{person}{Libo Ren}, \bibinfo{person}{Samuel Belkadi}, \bibinfo{person}{Lifeng Han}, \bibinfo{person}{Warren Del-Pinto}, {and} \bibinfo{person}{Goran Nenadic}.} \bibinfo{year}{2024}\natexlab{}.
\newblock \bibinfo{title}{Synthetic4Health: Generating Annotated Synthetic Clinical Letters}.
\newblock
\newblock
\showeprint[arxiv]{2409.09501}~[cs.CL]
\urldef\tempurl%
\url{https://arxiv.org/abs/2409.09501}
\showURL{%
\tempurl}


\bibitem[Sachan et~al\mbox{.}(2020)]%
        {sachan2020syntax}
\bibfield{author}{\bibinfo{person}{Devendra~Singh Sachan}, \bibinfo{person}{Yuhao Zhang}, \bibinfo{person}{Peng Qi}, {and} \bibinfo{person}{William Hamilton}.} \bibinfo{year}{2020}\natexlab{}.
\newblock \showarticletitle{Do syntax trees help pre-trained transformers extract information?}
\newblock \bibinfo{journal}{\emph{arXiv preprint arXiv:2008.09084}} (\bibinfo{year}{2020}).
\newblock


\bibitem[Sarawagi et~al\mbox{.}(2008)]%
        {sarawagi2008information}
\bibfield{author}{\bibinfo{person}{Sunita Sarawagi} {et~al\mbox{.}}} \bibinfo{year}{2008}\natexlab{}.
\newblock \showarticletitle{Information extraction}.
\newblock \bibinfo{journal}{\emph{Foundations and Trends{\textregistered} in Databases}} \bibinfo{volume}{1}, \bibinfo{number}{3} (\bibinfo{year}{2008}), \bibinfo{pages}{261--377}.
\newblock


\bibitem[Sharnagat(2014)]%
        {sharnagat2014named}
\bibfield{author}{\bibinfo{person}{Rahul Sharnagat}.} \bibinfo{year}{2014}\natexlab{}.
\newblock \showarticletitle{Named entity recognition: A literature survey}.
\newblock \bibinfo{journal}{\emph{Center For Indian Language Technology}} (\bibinfo{year}{2014}), \bibinfo{pages}{1--27}.
\newblock


\bibitem[Singh et~al\mbox{.}(2023)]%
        {singh2023unlocking}
\bibfield{author}{\bibinfo{person}{Shruti Singh}, \bibinfo{person}{Hitesh Lodwal}, \bibinfo{person}{Husain Malwat}, \bibinfo{person}{Rakesh Thakur}, {and} \bibinfo{person}{Mayank Singh}.} \bibinfo{year}{2023}\natexlab{}.
\newblock \showarticletitle{Unlocking Model Insights: A Dataset for Automated Model Card Generation}.
\newblock \bibinfo{journal}{\emph{arXiv preprint arXiv:2309.12616}} (\bibinfo{year}{2023}).
\newblock


\bibitem[Tian et~al\mbox{.}(2022)]%
        {tian2022improving}
\bibfield{author}{\bibinfo{person}{Yuanhe Tian}, \bibinfo{person}{Yan Song}, {and} \bibinfo{person}{Fei Xia}.} \bibinfo{year}{2022}\natexlab{}.
\newblock \showarticletitle{Improving relation extraction through syntax-induced pre-training with dependency masking}. In \bibinfo{booktitle}{\emph{Findings of the Association for Computational Linguistics: ACL 2022}}. \bibinfo{pages}{1875--1886}.
\newblock


\bibitem[Torfi et~al\mbox{.}(2020)]%
        {torfi2020natural}
\bibfield{author}{\bibinfo{person}{Amirsina Torfi}, \bibinfo{person}{Rouzbeh~A Shirvani}, \bibinfo{person}{Yaser Keneshloo}, \bibinfo{person}{Nader Tavaf}, {and} \bibinfo{person}{Edward~A Fox}.} \bibinfo{year}{2020}\natexlab{}.
\newblock \showarticletitle{Natural language processing advancements by deep learning: A survey}.
\newblock \bibinfo{journal}{\emph{arXiv preprint arXiv:2003.01200}} (\bibinfo{year}{2020}).
\newblock


\bibitem[Vaswani(2017)]%
        {vaswani2017attention}
\bibfield{author}{\bibinfo{person}{A Vaswani}.} \bibinfo{year}{2017}\natexlab{}.
\newblock \showarticletitle{Attention is all you need}.
\newblock \bibinfo{journal}{\emph{Advances in Neural Information Processing Systems}} (\bibinfo{year}{2017}).
\newblock


\bibitem[Yang et~al\mbox{.}(2024)]%
        {yang2024report}
\bibfield{author}{\bibinfo{person}{Blair Yang}, \bibinfo{person}{Fuyang Cui}, \bibinfo{person}{Keiran Paster}, \bibinfo{person}{Jimmy Ba}, \bibinfo{person}{Pashootan Vaezipoor}, \bibinfo{person}{Silviu Pitis}, {and} \bibinfo{person}{Michael~R Zhang}.} \bibinfo{year}{2024}\natexlab{}.
\newblock \showarticletitle{Report Cards: Qualitative Evaluation of Language Models Using Natural Language Summaries}.
\newblock \bibinfo{journal}{\emph{arXiv preprint arXiv:2409.00844}} (\bibinfo{year}{2024}).
\newblock


\bibitem[Young et~al\mbox{.}(2018)]%
        {8416973}
\bibfield{author}{\bibinfo{person}{Tom Young}, \bibinfo{person}{Devamanyu Hazarika}, \bibinfo{person}{Soujanya Poria}, {and} \bibinfo{person}{Erik Cambria}.} \bibinfo{year}{2018}\natexlab{}.
\newblock \showarticletitle{Recent Trends in Deep Learning Based Natural Language Processing [Review Article]}.
\newblock \bibinfo{journal}{\emph{IEEE Computational Intelligence Magazine}} \bibinfo{volume}{13}, \bibinfo{number}{3} (\bibinfo{year}{2018}), \bibinfo{pages}{55--75}.
\newblock
\urldef\tempurl%
\url{https://doi.org/10.1109/MCI.2018.2840738}
\showDOI{\tempurl}


\bibitem[Yuan et~al\mbox{.}(2019)]%
        {Yuan-etal-2019-distant}
\bibfield{author}{\bibinfo{person}{Changsen Yuan}, \bibinfo{person}{Heyan Huang}, \bibinfo{person}{Chong Feng}, \bibinfo{person}{Xiao Liu}, {and} \bibinfo{person}{Xiaochi Wei}.} \bibinfo{year}{2019}\natexlab{}.
\newblock \showarticletitle{Distant supervision for relation extraction with linear attenuation simulation and non-IID relevance embedding}. In \bibinfo{booktitle}{\emph{Proceedings of the Thirty-Third AAAI Conference on Artificial Intelligence and Thirty-First Innovative Applications of Artificial Intelligence Conference and Ninth AAAI Symposium on Educational Advances in Artificial Intelligence}} (Honolulu, Hawaii, USA) \emph{(\bibinfo{series}{AAAI'19/IAAI'19/EAAI'19})}. \bibinfo{publisher}{AAAI Press}, Article \bibinfo{articleno}{911}, \bibinfo{numpages}{8}~pages.
\newblock
\showISBNx{978-1-57735-809-1}
\urldef\tempurl%
\url{https://doi.org/10.1609/aaai.v33i01.33017418}
\showDOI{\tempurl}


\bibitem[Zeng et~al\mbox{.}(2014)]%
        {zeng2014relation}
\bibfield{author}{\bibinfo{person}{Daojian Zeng}, \bibinfo{person}{Kang Liu}, \bibinfo{person}{Siwei Lai}, \bibinfo{person}{Guangyou Zhou}, {and} \bibinfo{person}{Jun Zhao}.} \bibinfo{year}{2014}\natexlab{}.
\newblock \showarticletitle{Relation classification via convolutional deep neural network}. In \bibinfo{booktitle}{\emph{Proceedings of COLING 2014, the 25th international conference on computational linguistics: technical papers}}. \bibinfo{pages}{2335--2344}.
\newblock


\bibitem[Zhang et~al\mbox{.}(2009)]%
        {zhang2009rule}
\bibfield{author}{\bibinfo{person}{Chunju Zhang}, \bibinfo{person}{Xueying Zhang}, \bibinfo{person}{Wenming Jiang}, \bibinfo{person}{Qijun Shen}, {and} \bibinfo{person}{Shanqi Zhang}.} \bibinfo{year}{2009}\natexlab{}.
\newblock \showarticletitle{Rule-based extraction of spatial relations in natural language text}. In \bibinfo{booktitle}{\emph{2009 international conference on computational intelligence and software engineering}}. IEEE, \bibinfo{pages}{1--4}.
\newblock


\bibitem[Zhang and Wang(2015)]%
        {zhang2015relationclassificationrecurrentneural}
\bibfield{author}{\bibinfo{person}{Dongxu Zhang} {and} \bibinfo{person}{Dong Wang}.} \bibinfo{year}{2015}\natexlab{}.
\newblock \bibinfo{title}{Relation Classification via Recurrent Neural Network}.
\newblock
\newblock
\showeprint[arxiv]{1508.01006}~[cs.CL]
\urldef\tempurl%
\url{https://arxiv.org/abs/1508.01006}
\showURL{%
\tempurl}


\end{thebibliography}

\end{document}